This article has been accepted for publication in IEEE Intelligent Systems. This is the author's version which has not been fully edited and content may change prior to final publication. Citation information: DOI 10.1109/MIS.2025.3588759

# Lightweight Attribute Localizing Models for Pedestrian Attribute Recognition

Ashish Jha, *Skolkovo Institute of Science and Technology (SKOLTECH), Russia*
Dimitrii Ermilov, *Skolkovo Institute of Science and Technology (SKOLTECH), Russia*
Anh Huy Phan, *Skolkovo Institute of Science and Technology (SKOLTECH), Russia*
Konstantin Sobolev, *Skolkovo Institute of Science and Technology (SKOLTECH), Russia*
Salman Ahmadi-Asl, *SKOLTECH, and Innopolis University, Russia*
Naveed Ahmed, *University of Sharjah, UAE*
Imran Junejo, *Advanced Micro Devices (AMD), Canada*
Zaher AL Aghbari, *University of Sharjah, UAE*
Thar Baker, *University of Brighton, UK*
Ahmed Mohamed Khedr, *University of Sharjah, UAE*
Andrzej Cichocki, *Skolkovo Institute of Science and Technology (SKOLTECH), Russia*

*Abstract—Pedestrian Attribute Recognition (PAR) focuses on identifying various attributes in pedestrian images, with key applications in person retrieval, suspect re-identification, and soft biometrics. However, Deep Neural Networks (DNNs) for PAR often suffer from over-parameterization and high computational complexity, making them unsuitable for resource-constrained devices. Traditional tensor-based compression methods typically factorize layers without adequately preserving the gradient direction during compression, leading to inefficient compression and a significant accuracy loss. In this work, we propose a novel approach for determining the optimal ranks of low-rank layers, ensuring that the gradient direction of the compressed model closely aligns with that of the original model. This means that the compressed model effectively preserves the update direction of the full model, enabling more efficient compression for PAR tasks. The proposed procedure optimizes the compression ranks for each layer within the ALM model, followed by compression using CPD-EPC or truncated SVD. This results in a reduction in model complexity while maintaining high performance.*

Convolutional Neural Networks (CNNs) have proven to be effective for various computer vision tasks, such as image recognition, object detection, and pose estimation. However, many recent CNNs face challenges due to high computational complexity and over-parameterization, which hinder their deployment on resource-constrained devices like smartphones and surveillance cameras. Reducing the parameters and computational cost of neural networks is a critical research area, although achieving significant reduction without sacrificing inference accuracy remains a challenge. Four main approaches for model reduction have emerged, low-rank tensor approximation, pruning, quantization, and knowledge distillation. Pruning algorithms, while effective, are computationally expensive and struggle with runtime speed-up or changes in network structure. Quantization, though promising, often leads to issues with backpropagation and convergence, significantly impacting accuracy. Knowledge distillation transfers knowledge from a larger model to a smaller one, but discrepancies often remain between the predictive distributions of the models, limiting its effectiveness.

In this work we focus on reducing network layers through tensor decomposition, addressing the issue of preserving the gradient direction during compression. We propose a novel rank selection method to preserve the gradient direction, followed by tensor decomposition to create Light-Weight (LW) layers. Convolutional weights are known to be in the low-rank space and can be reduced using this approach with minimal loss in accuracy. Unlike previous methods such as [1] and [2] which typically focus on compressing specific layers without gradient preservation and exhibit higher accuracy losses, our approach improves accuracy drop during fine-tuning and further improves the reduction in both computational cost and parameters.

In this work we have combined tensor decomposition algorithms to compress all layers [3] with preserved gradient direction such that the gradient direction of the compressed model aligns closely with that of the original model. This procedure optimizes compression ranks for each layer, followed by CPD-EPC [2] or truncated SVD [4] compression. This results in reduced complexity while maintaining performance, as demonstrated by our experiments in Section 6.

Rank determination remains a challenging problem in tensor decomposition and its application in model compression. Existing approaches such as [2] often rely on binary search to identify suitable ranks for low-rank layers, this process requires finetuning the entire neural network at each rank until a stable accuracy is achieved before determining the next suitable rank. Moreover, existing methods do not preserve the gradient direction during decomposition resulting in less precise gradients and acute accuracy loss which cannot be completely recovered using fine-tuning.

Working on this, we propose a compression strategy which ensures the low-rank layers preserve primary direction of the gradients from the original dense layers allowing performance recovery during fine-tuning, Additionally, we constrain the norms of the factorized layers during fine-tuning, which helps mitigate over-fitting caused by variance introduced during the decomposition process.

The main contributions of this paper are summarized as follows

- We propose a compression strategy ensuring the low-rank layers preserves the primary direction of the gradients from the original dense layers, followed by compression using CPD-EPC and truncated SVD, resulting in the LWALM.
- Constraints on the factorized layers during fine-tuning, mitigating over-fitting.

The rest of the paper is structured as follows: Section 1 reviews recent advancements in NN compression. Section 2 introduces the notations and concepts used. Section 3 discusses PAR problems and algorithms, including the ALM. Section 4 details the proposed compression algorithm. Section 5 explains the compression of ALM layers to obtain LWALM. Section 6 presents the results. Section 7 evaluates LWALMs using confidence calibration, and Section 8 concludes with future research directions.

## 1. Related Works

The problem of neural network (NN) compression has been extensively studied to address overparameterization and computational inefficiency. Early work by Denil et al. [5] introduced redundancy reduction, demonstrating that a significant portion of network parameters could be predicted from a smaller subset. This opened the door to compression techniques like the truncated Singular Value Decomposition (SVD) applied by Denton et al. [6] to fully connected layers, achieving compression with minimal accuracy loss. However, their method focused mainly on specific layers rather than the entire network. Similarly, Lebedev et al. [7] proposed low-rank approximations for convolutional layers but limited their approach to a few layers, which constrained broader model-wide applications. Subsequent methods sought to expand the scope of compression. Vector quantization [6] and tensor train decomposition [8] offered promising results, though these methods often lacked a robust mechanism for determining optimal compression ranks across layers. Techniques such as PCA-based rank selection and pruning by Han et al. [9] reduced parameters network-wide but did not fully address the complexity of layer-wise rank determination, crucial for achieving fine-grained compression balance. Additionally, practical methods like FFT-based convolution speedups [10] and CPU code optimizations [11] focused on computational efficiency but were less effective in reducing overall model size.

More recently, tensor decomposition approaches have shown potential for comprehensive NN compression. Yin et al. [12] proposed an optimization framework leveraging CPD and Tucker decompositions to balance compression and model performance, making them suitable for resource-constrained environments. Similarly, Astrid and Lee [13] employed CP decomposition with the Tensor Power Method, compressing convolutional layers while maintaining accuracy. However, these methods primarily focus on static rank selection strategies, which may lead to suboptimal compression in complex models.

In contrast, our work introduces a novel dynamic rank search procedure that optimally determines layer-wise compression ranks, addressing the limitations of previous static approaches. Unlike earlier works that applied a single decomposition technique across the model, we integrate CPD-EPC for layers with larger kernels and truncated SVD for smaller kernels, providing a tailored, adaptive compression strategy. Furthermore, we enhance fine-tuning with a modified L2-norm-constrained loss function to mitigate overfitting and maintain model complexity, a significant improvement over existing methods that do not explicitly address overfitting post-compression. This comprehensive approach results in the Lightweight Attribute Localization Model (LWALM), capable of achieving high compression rates without sacrificing performance, thus extending the applicability of PAR models to low-resource environments.

## 2. Preliminary notations and concepts

This section presents basic definitions and concepts used throughout the paper. Tensors are denoted by underlined bold capital letter, e.g. $\underline{\mathbf{X}}$, matrices by bold capital letters, e.g. $\mathbf{X}$, and vectors by lower case letters, e.g. $\mathbf{x}$.

### 2.1. Canonical Polyadic Decomposition with Error Preserving Correction (CPD-EPC)

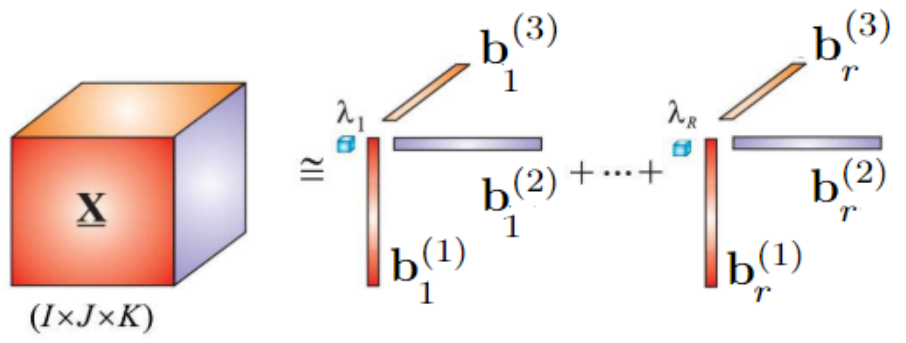

**FIGURE 1.** Rank-R CP decomposition [14]

CPD approximates an $N^{th}$-order tensor, $\underline{\mathbf{X}} \in \mathbb{R}^{I_1 \times I_2 \times \cdots \times I_N}$ as a sum of rank-1 tensors as illustrated in Fig. 1

$$\underline{\mathbf{X}} \cong \underline{\mathbf{Y}} = \sum_{r=1}^{R} \lambda_r \, \mathbf{b}_r^{(1)} \circ \mathbf{b}_r^{(2)} \circ \cdots \circ \mathbf{b}_r^{(N)}$$

where $\lambda_r$ are the rank-1 tensor weights, "$\circ$" denotes the tensor outer product. During CPD computation with a given tensor rank, some instability issues may occur; e.g., when the rank exceeds the tensor dimension, or loading components are highly collinear in several modes, or CPD has no optimal solution. This instability causes norms of rank-1 terms to become significantly large but cancel each other. CPD-EPC corrects the unstable decomposition by minimizing norms of these high norm rank-1 tensors obtained during the CPD computation [2]. More precisely, a new tensor $\underline{\mathbf{Y}}$ with minimum rank-1 tensor norms is found while it still explains $\underline{\mathbf{X}}$ at the current approximation error level. Such decomposition improves stability and convergence and can be achieved by solving the constrained CPD approximation given by

$$\min \ f(\theta) = \sum_{\mathbf{r}=\mathbf{1}}^{\mathbf{R}} \lambda_{\mathbf{r}}^{\mathbf{2}}, \quad \text{s.t.} \ \mathbf{c}(\theta) = \|\underline{\mathbf{X}} - \underline{\mathbf{Y}}\|_{\mathbf{F}}^{\mathbf{2}} \leq \delta^{\mathbf{2}},$$

where $\theta$ is a vector of all parameters of the decomposition, $\delta$ is a approximation error bound, and the weights $\lambda_r^2$ represent the Frobenius norm of the $r^{th}$ rank-1 tensor given that loading components $\boldsymbol{u}_r^{(n)}$ are unit-length vectors, $\left\| \lambda_r \boldsymbol{b}_r^{(1)} \circ \boldsymbol{b}_r^{(2)} \circ \cdots \circ \boldsymbol{b}_r^{(N)} \right\|_F^2 = \lambda_r^2$. This method is called Error Preserving Correction (EPC) method [2].

## 3. PAR and ALM

Pedestrian Attribute Recognition (PAR) recognizes pedestrian attributes from a target image. It has found a growing interest in the computer vision community due to its applications in video surveillance. More precisely, given an input image $X$, and several predefined attributes $N$ as $X = \{I_1, I_2, I_3, \dots, I_N\}$, the goal is to predict attributes of a pedestrian in the image $X$ such as *bag, hair, shoes, etc*. The holistic methods such as DeepSAR and DeepMAR [7] consider PAR a multi-label classification problem and rely on global feature representations. Attribute Localization Model (ALM) addresses the attribute recognition by enhancing a deeper understanding of attributes through localization mechanisms. In [3], Tang et.al propose a Deep Attribute Localization Model (ALM) which can automatically discover discriminative regions and extract region-based feature representations in a pedestrian image. Their model gives a state-of-the-art classification accuracy for PAR metrics. In the following sections, we explain algorithms for creating and fine-tuning the Lightweight Attribute Localization Model by compressing the components of ALM.

## 4. Proposed CNN Compression using CPD-EPC and SVD

This section outlines the methodology proposed for compressing the model. The Attribute Localization Model (ALM) is built on top of the BN-Inception backbone and follows a feature pyramid structure, as presented in [3]. To compress entire model, each convolutional layer is replaced by a sequence of lightweight convolutions. The rank of each layer is first determined to identify the optimal compression, and the layer is then decomposed accordingly. The detailed process of this methodology is discussed in the following subsections.

### 4.1. Compression using CPD-EPC

In the ALM, we use CPD-EPC for convolutional kernels with large kernel size (e.g., 3, 5) at two different error bounds ($\delta = 0.001$ and $\delta = 0.002$) and SVD for $1 \times 1$ convolutional kernels. The obtained LWALM is compared to the results of other compression techniques, such as Tucker-2 and CP. For a convolutional tensor of size $D \times D \times S \times T$, the $1^{st}$ and $2^{nd}$ modes ($D \times D$) are the spatial width and height, while the $3^{rd}$ and $4^{th}$ modes ($S$ and $T$) are the input and output channels, respectively. For a given tensor rank $R$, and an approximation error bound, we perform the following steps.

In the first step, the tensor is fitted by a standard CP model to replace convolution kernel $\underline{\mathbf{K}}$ with three consecutive convolutions. It is convenient to represent $4^{th}$ order tensor $D \times D \times S \times T$ as $3^{rd}$ order tensor to allow more balanced dimensions and less layers after decomposition; therefore, it is reshaped into the size of $D^2 \times S \times T$. The rank-$R$ CP-decomposition of the $3^d$ order tensor has the form

$$\underline{\mathbf{K}}(t, s, \{j, i\}) \cong \sum_{r=1}^{R} \underline{\mathbf{T}}^{hw}(\{j, i\}, r)\, \underline{\mathbf{T}}(s, r)\, \underline{\mathbf{T}}^{t}(t, r), \quad (1)$$

where $\underline{\mathbf{T}}^{hw}(\{j, i\}, r)$, $\underline{\mathbf{T}}(s, r)$, and $\underline{\mathbf{T}}^{t}(t, r)$ are of sizes $(D^2 \times R)$, $S \times R$, and $T \times R$, respectively. Therefore, $1 \times 1$ convolution projects input from $C_{in}$ input channels into $R$ channels. Then, group convolution layer applies $R$ separate convolutions with kernel size $3 \times 3$ or $7 \times 7$, one for each channel. Finally, one more $1 \times 1$ convolution expands $R$ channels into $C_{out}$ channels to get the output.

In the second step, we evaluate whether the norm $\|\boldsymbol{\lambda}_{[k]}\|_2^2$ of the rank-1 tensor components obtained in Step 1 exceeds a predefined threshold $\gamma^2$. Here, $\boldsymbol{\lambda}_{[k]}$ is the vector of weights of all rank-1 tensors at the $k$-th iteration, defined as

$$\boldsymbol{\lambda}_{[k]} = [\lambda_1^{(k)}, \lambda_2^{(k)}, \dots, \lambda_R^{(k)}],$$

where $R$ is the rank of the tensor decomposition. If $\|\boldsymbol{\lambda}_{[k]}\|_2^2 \geq \gamma^2$, the resulting tensor is considered unstable. To address this, a correction method is applied to $\mathbf{Y}_{[k]}$, producing a new tensor $\mathbf{Y}_{[k+1]}$ with a minimum norm $\|\boldsymbol{\lambda}_{[k+1]}k + 1\|_2^2$, while ensuring

$$\|\mathbf{X} - \mathbf{Y}_{[k+1]}\|_F \leq \|\mathbf{X} - \mathbf{Y}_{[k]}\|_F.$$

If the norm condition is not exceeded, the standard CP decomposition from Step 1 is applied to $\mathbf{Y}$ to compute $\mathbf{Y}_{[k+1]}$ using $\mathbf{X}_{[k]}$. The resulting tensor serves as a feasible solution, with the current approximation error

$$\delta = \|\mathbf{Y} - \mathbf{X}_{[k-1]}\|_F,$$

as described in [2].

### 4.2. Compression using SVD

SVD is the matrix decomposition that represents matrix in the form of $\mathbf{X} = \mathbf{USV}^\top$, where $\mathbf{S}$ is diagonal matrix while $\mathbf{U}$ and $\mathbf{V}$ are unitary matrices [15]. In case of $1 \times 1$ convolution, the weight is matrix of size $C_{in} \times C_{out}$. Therefore, it can be replaced with two $1 \times 1$ convolutions where $1^{st}$ convolution projects $C_{in}$ input channels into $R$ channels and $2^{nd}$ one expands $R$ channels into $C_{out}$ output channels, with weights $\mathbf{V}^\top$ and $\mathbf{US}$, respectively.

## 5. Compression Approach

Convolution operations contribute to the bulk of computations in ALM [3] and the model is embedded with Attention Modules (AM) at different levels of the Inception-V2 backbone (incep-3b,4d,5d). The number of embedded AM's is dependent on the number of attributes in the training dataset; e.g., $35 \times 3 = 105$ for PETA with 35 attributes and 3 different levels of Inception-V2 backbone. Each AM consists of two Conv2D layers with shapes: (768, 48) and (48, 768) for incep_3b, (512, 32) and (32, 512) for incep_4d, and (256, 16) and (16, 256) for incep_5b; thus, there are a total of $105 \times 2 = 210$ ($1 \times 1$) convolutional layers for all AMs. Additionally, 31 layers have kernels with size $3 \times 3$ and a single layer has kernel with size $7 \times 7$. There are 17.1 million parameters in ALM.

We apply a combination of two aforementioned approaches to the ALM. Firstly, each standard convolutional kernel is factorized with the optimal rank $R$ to obtain a sequence of LW layers. Then, we replace the former layer in ALM with latter LW layers and fine-tune the model on PAR datasets.

### 5.1. Rank Search Procedure

A key challenge in tensor decomposition based NN compression is determining the optimal rank for decomposition. This rank, denoted as $R$, must balance compression with the potential loss of accuracy. In this section, we first begin by reviewing the binary search method for rank selection and then introduce a novel gradient matching based approach for optimal rank selection.

*5.1.1. Binary Search.* In an iterative heuristic binary search based rank search [16], the smallest acceptable rank $R$ for each layer is found. The process begins by identifying the maximum possible rank for the decomposition of the weight tensor at each layer, Then, the binary search algorithm is applied iteratively to factorize each layer, carefully monitoring the impact of accuracy loss at each rank. Fine-tuning after each decomposition ensures that the drop in accuracy does not exceed a predefined threshold.

*5.1.2. Rank-Search Procedure Using Gradient Alignment.* Instead of directly measuring the impact of accuracy loss at each rank, we propose selecting $\mathcal{R}$ such that the gradient direction of the compressed model closely aligns with the gradient direction of the full model during finetuning. This procedure identifies the optimal low-rank approximation for convolutional layers by minimizing the distance between the gradients of the low-rank model and a reference gradient from the original network. The process is applied sequentially, starting from the first convolutional layer and progressing to the last.

To begin, the original (uncompressed) network is fine-tuned for one epoch. The gradient of the last layer with respect to the loss function is computed and stored as $\mathbf{g}_{\text{ref}} = \nabla_\Theta L(\Theta; \mathcal{X})$ serving as the reference gradient. Next, a set of candidate ranks, $\{R_1, R_2, \dots, R_m\}$, is defined to explore potential low-rank approximations. For the first convolutional layer, the kernel is decomposed using each candidate rank $R_j$. After decomposing with rank $R_j$, the low-rank model is fine-tuned for one epoch, and the gradient of the last layer is computed as $\mathbf{g}_j$. The distance between $\mathbf{g}_j$ and the reference gradient $\mathbf{g}_{\text{ref}}$ is then measured using the $\ell_2$-norm. The rank $R^*$ that minimizes this distance is selected as the optimal rank for the layer

$$R^* = \arg\min_{R_j} \quad \|\mathbf{g}_{\text{ref}} - \frac{\mathbf{g}_j^T \mathbf{g}_{\text{ref}}}{\mathbf{g}_j^T \mathbf{g}_j} \mathbf{g}_j\|_2.$$

Here, the term $\frac{\mathbf{g}_j^T \mathbf{g}_{\text{ref}}}{\mathbf{g}_j^T \mathbf{g}_j}$ ensures that the gradient $\mathbf{g}_j$ is scaled to align with $\mathbf{g}_{\text{ref}}$ in order to mitigate the effects of differences in magnitude. This procedure is repeated for subsequent convolutional layers. At each step, the already decomposed layers are retained with the optimal ranks, and the gradient computation incorporates the cumulative effect of previous rank selections. By ensuring that the selected rank minimizes deviation from the original gradient flow, this approach maintains alignment with the fine-tuning objective and reduces performance loss due to low-rank approximations.

## 5.2. Ablation study

In this section, we present an ablation study to evaluate the impact of our proposed compression method on computational efficiency across various layers of ALM. By analyzing the speedup achieved through different kernel sizes, we aim to understand how different techniques contribute to reducing computational complexity. Table 1 presents a comparison of the speedup achieved by our method across different layers on 3 kernel sizes ($1 \times 1, 3 \times 3, 7 \times 7$). The speedup is calculated as the ratio of the GFLOPs of the ALM layer to the sum of GFLOPs from each decomposed LWALM layer

$$\text{Speedup} = \frac{\text{GFLOPs}(\text{Layer}_{\text{ALM}})}{\sum_i \text{GFLOPs}(\text{Layer}^i_{\text{LWALM}})}.$$

Our results suggest that small to medium kernel sizes, such as $1 \times 1$ and $3 \times 3$, lead to more substantial speedups after compression. For instance, at the 3*a*_3*x*3 and conv2_3x3 (medium-sized) layers, speedups of 14.77x and 6.1x are achieved, respectively, highlighting the significant efficiency gains from CPD-EPC in medium-sized kernel layers. Additionally, SVD decomposition effectively reduces the complexity of small to medium-sized layers; for example, at the 3a_3x3_reduce (small-sized) layer, SVD achieves a 3.4x reduction. In contrast, smaller gains are observed for certain medium-sized layers, such as a 1.2x reduction for inception_3a_3x3 using SVD, and for large-sized layers like conv1_7x7_s2, where CPD-EPC yields only a 1.1x speedup. Moreover, we did not observe any practical benefit from applying SVD to large layers such as conv1_7x7_s2. These findings indicate that CPD-EPC is most effective for medium to high-dimensional layers, while SVD is better suited for small to intermediate-sized layers, emphasizing the need for tailored decomposition strategies based on kernel size and layer characteristics.

# 6. Experiments

This section presents the results of compressing the ALM, focusing on improvements in speed and computational efficiency. Additionally, we evaluate LWALMs for the PAR task using two popular datasets, PA-100K and PETA, comparing them against two benchmark compression methods Tucker-2 and CPD, as well as other state-of-the-art (SOTA) methods for PAR. The benchmark models were created following a similar approach described in Section 5, replacing CPD-EPC with CPD and Tucker to generate the respective models.

**TABLE 1.** Comparison of Speedup Across Layers

| Algorithm | Per layer GFLOPs LWALM | ALM | Speedup |
|---|---|---|---|
| *conv2_3x3 [Kernel: 3x3]* | | | |
| CPD-EPC | L0=2.63e-2<br>L1=3.30e-3<br>L2=6.60e-2 | 1.62e-1 | **6.1**× |
| *3a_3x3_reduce [Kernel: 1x1]* | | | |
| SVD | L0=1.33e-3<br>L1=4.92e-4 | 4.50e-3 | **3.4**× |
| *3a_3x3 [Kernel: 3x3]* | | | |
| CPD-EPC | L0 = 3.846e-4<br>L1 = 5.63e-5<br>L2 = 4.702e-4 | 1.346e-2 | **14.77** |
| *inception_3a_3x3 [Kernel: 3x3]* | | | |
| SVD | L0=0.0277<br>L1=0.0031 | 0.037 | **1.2**× |
| *conv1_7x7_s2 [Kernel: 7x7]* | | | |
| CPD-EPC | L0=0.014<br>L1=0.057<br>L2=0.075 | 0.155 | **1.1**× |
| SVD | L0=0.1541<br>L1=0.0676 | 0.155 | **0.69**× |

## 6.1. Losses

During the model compression stage, the norms of rank-1 tensors are explicitly minimized. However, during fine-tuning, these norms can grow excessively, which may lead to instability in training and hinder convergence. To address this, we introduce an additional regularization term that penalizes the loss function whenever the norms of the rank-1 tensors become too large during fine-tuning. This regularization helps control the growth of the tensor norms, reducing model complexity, and improving generalization. It also mitigates the risk of overfitting by discouraging the model from becoming overly dependent on specific features that may not generalize well. Formally, we minimize the following objective function

$$\text{minimize} \quad \mathcal{L}_{LWALM} + \lambda \sum_{L=1}^{N} \sum_{l=1}^{n} \left\| D_l^{(L)} \right\|_F^2 \qquad (2)$$

where, $D_l^{(L)}$ is the $l^{th}$ factorized layer for the corresponding ALM layer $L$. $n$ is the number of factorized layers i.e. for CPD $n = 3$ and for SVD $n = 2$.

$\mathcal{L}_{LWALM}$ stands for the weighted binary cross-entropy loss, $L$ is a set of $N$ ALM layers on which compression was performed and $\lambda$ is the shrinkage factor.

## 6.2. Training

LWALMs were trained on a Tesla-T4 GPU with 26 GB memory in two batches of sizes 32 and 64. The initial learning rate was set to 0.001 with an adjustment of 0.1× after every 10 epochs. Adam optimizer with a weight decay of 0.0005 and the proposed loss function with $\lambda = 0.001$ were used during fine-tuning/training.

## 6.3. Datasets

Two well-known PAR datasets, PETA [17] and PA-100K [18], were used for evaluation. To make a fair comparison with the ALM, we used the same data partitions for both datasets as mentioned in their work [3]. PETA was evaluated at each attribute's mean recognition accuracy, which is given by the average of positive and negative examples' recognition accuracy. The widely used evaluation method is a random dataset division into multiple parts, 9500 for training, 1900 for verifying and 7600 for testing [7]. Similarly, for PA-100K the entire dataset was randomly split into 80,000 training images, 10,000 validation images, and 10,000 test images.

## 6.4. Evaluation Metric

We evaluated the models using standard metrics such as mean accuracy(mA), F1 score, Recall, and precision (Prec), commonly used in the PAR task. Additionally, we used the Performance per Parameter Count (PPP), defined as PPP = $\frac{\text{Performance}}{\text{Parameter Count}}$, to assess the model's efficiency.

## 6.5. Performance Comparison

We compare LWALMs with PAR models in 4 different categories: (1) Holistic methods, including ACN and DeepMar, (2) Relation-based methods, (3) Attention-based methods, and (4) Part-based methods. Table 2 provides a detailed comparison of Lightweight Attribute Learning Models (LWALMs) against other state-of-the-art PAR models, as well as alternative methods such as Tucker decomposition (TKD) [19] and CANDECOMP/PARAFAC (CPD) [20]. LWALMs consistently demonstrate remarkable performances, particularly when evaluated using the Parameter-Performance Ratio (PPP) metric.

On the PETA dataset, LWALM ($\delta = 0.002$) achieves a PPP of 12.50 with only 0.68 GFLOPs. In comparison, (TKD+SVD) delivers almost the same PPP score of 12.48, but at the cost of 1.37 GFLOPs and a reduction in accuracy of over 3%. Similarly, (CPD+SVD) achieves a PPP score of 8.29 and requires 1.66 GFLOPs. Despite this, LWALMs maintain high F1 scores, with values of 85.64 for $\delta = 0.002$ and 85.80 for $\delta = 0.001$, outpacing (TKD+SVD) at 83.78 and (CPD+SVD) at 81.88. On the PA-100K dataset, LWALMs ($\delta = 0.002$) achieve an impressive PPP of 66.00 with just 0.23 GFLOPs, outperforming (TKD+SVD) at 16.72 PPP with 1.24 GFLOPs an improvement of over 3×, and (CPD+SVD) at 7.54 PPP with 1.73 GFLOPs. Additionally, LWALMs maintain competitive F1 scores, achieving 85.65 and 86.27, compared to (TKD+SVD) at 85.24

**TABLE 2.** Performance Comparison between LWALMs with PAR Models

| | PETA | | | | | | | | PA-100K | | | | | | | |
|---|---|---|---|---|---|---|---|---|---|---|---|---|---|---|---|---|
| | *PAR Models* | | | | | | | | | | | | | | | |
| Models | #P | GFLOPs | mA | Accu | Pree | Recall | f1 | PPP | #P | GFLOPs | mA | Accu | Pree | Recall | f1 | PPP |
| DeepMar | 58.5M | - | 82.89 | 75.07 | 83.68 | 83.14 | 83.41 | 1.42 | 58.5M | - | 72.7 | 70.39 | 82.24 | 80.42 | 81.32 | 1.24 |
| GRL | >50M | $>10$ | 86.7 | - | 84.34 | 88.82 | 86.51 | 1.73 | >50M | $>10$ | - | - | - | - | - | - |
| VeSPA | 17.0M | $>3$ | 83.45 | 77.73 | 86.18 | 84.81 | 85.49 | 4.91 | 17.0M | $>3$ | 76.32 | 73.00 | 84.99 | 81.49 | 83.2 | 4.49 |
| PGDM | 87.2M | | 82.97 | 78.08 | 86.86 | 84.68 | 85.76 | 0.95 | 87.2M | | 74.95 | 73.08 | 84.36 | 82.24 | 83.29 | 0.86 |
| LG-Net | >20M | $>4$ | - | - | - | - | - | - | >20M | $>4$ | 76.96 | 75.55 | 86.99 | 83.17 | 85.04 | 3.85 |
| BN-Inception | 10.3M | 1.78 | 82.66 | 77.73 | 86.68 | 84.2 | 85.57 | 8.02 | 10.3M | 1.78 | 77.47 | 75.05 | 86.61 | 85.34 | 85.97 | 7.52 |
| $C^2T$-Net | 389M | 6.5 | 88.20 | - | - | - | **91.00** | 0.23 | 389M | 6.5 | **87.20** | - | - | - | **89.10** | 0.22 |
| UniHCP | 109.1M | 1.76 | **88.78** | - | - | - | - | 0.81 | 109.1M | 1.76 | 86.18 | - | - | - | - | 0.79 |
| ALM | 17.1M | 1.95 | 86.30 | 79.52 | 85.65 | 88.09 | 86.85 | 5.05 | 14.02M | 1.95 | 80.68 | 77.08 | 84.21 | 88.84 | 86.46 | 5.75 |
| *LWALM ($\delta = 0.002$) | 6.72M | **0.68** | 84.01 | 76.21 | 83.21 | 88.22 | 85.64 | **12.50** | **1.2M** | **0.23** | 79.2 | 76.42 | 83.21 | 88.25 | 85.65 | **66.00** |
| *LWALM ($\delta = 0.001$) | 7.28M | 0.93 | 85.63 | 77.04 | 83.76 | 87.96 | 85.80 | 11.76 | 1.6M | 0.80 | 80.24 | 77.04 | 84.21 | 88.45 | 86.27 | 50.15 |
| | *Compression Methods* | | | | | | | | | | | | | | | |
| TKD + SVD | **6.61M** | 1.37 | 82.51 | 74.5 | 80.14 | 87.77 | 83.78 | 12.48 | 4.71M | 1.24 | 78.77 | 75.41 | 82.8 | 87.83 | 85.24 | 16.72 |
| CPD + SVD | 9.7M | 1.66 | 80.38 | 72.13 | 77.31 | 87.03 | 81.88 | 8.29 | 10.6M | 1.73 | 79.91 | 77.01 | 83.9 | 88.77 | 86.27 | 7.54 |

and (CPD+SVD) at 86.27.

Furthermore, our LWALMs show strong performance in multiple metrics. For example, on the PETA dataset, LWALMs ($\delta = 0.001$) achieve a recall of 87.96, closely matching that of the larger models while requiring only a fraction of the resources (Table 2).

The heavyweight PAR models such as $C^2$T-Net and UniHCP achieve slightly higher accuracy but have much lower PPP values due to their immense resource demands. For instance, $C^2$T-Net has a PPP of only 0.23 with 6.5 GFLOPs, and UniHCP achieves a PPP of 0.81 with 1.76 GFLOPs. In contrast, LWALMs offer a vastly superior balance between accuracy and resource efficiency, as evidenced by their high PPP scores on both datasets. These results highlight LWALMs' suitability for edge devices and embedded systems.

## 7. Confidence Calibration

Calibration is the problem of estimating probability which represents the likelihood of actual correctness and has recently shown its importance in modern neural network models [21]. It is widely used in applications where decisions rely on predicted probabilities. We employ a binary classification approach for confidence calibration.

### 7.1. Reliability Diagrams

Reliability diagrams are tools to visualize model calibration [21]. A diagonal line represents the identity function, and any deviation from the diagonal represents miscalibration. These diagrams are plotted as accuracy with confidence [21]. All uncalibrated predictions $\widehat{p}_i$ corresponding to $l$ attributes ($i = 1, 2, \ldots, l$) for all samples are divided into the mutually exclusive bins $B_1, B_2, \ldots, B_M$, and the bin boundaries $0 = a_1 \leq a_2 \leq \cdots \leq a_{M+1} = 1$, where the bin $B_m$ is defined as the interval $(a_m, a_{m+1}]$. Given fixed bin boundaries for the bin $B_m$ with interval $(a_m, a_{m+1}]$ consisting of positive-class samples and predictions $\widehat{p}_i$ in the interval $(a_m, a_{m+1}]$, their accuracy and confidence can be calculated as follows

$$Acc(B_m) = (\sum_{p_i \in B_m} p_i) \,/\, |\{\widehat{p}_i | a_m \leq \widehat{p}_i < a_{m+1}\}|, \quad (3)$$

$$Conf(B_m) = (\sum_{\widehat{p}_i \in B_m} \widehat{p}_i) \,/\, |\{\widehat{p}_i | a_m \leq \widehat{p}_i < a_{m+1}\}|, \quad (4)$$

where, $p_i$, $i = 1, 2 \ldots, l$ are the true labels.

### 7.2. Expected Calibration Error (ECE)

ECE is a scalar representation for model calibration and is approximated by first partitioning uncalibrated predictions $\widehat{p}_i$ and its corresponding true labels $p_i$ into $M$ bins. Then a weighted average of the bins accuracy/confidence difference is taken as follows:

$$ECE = \sum_{m=1}^{M} \left( \frac{|a_m \leq \widehat{p}_i < a_{m+1}|}{n} \right) (|Acc(B_m) - Conf(B_m)|),$$

where $n$ is the number of samples and the difference between Acc(uracy) and Conf(idence) on the right-hand side gives *calibration gap*. [21].

### 7.3. Temperature Scaling

For temperature scaling, a calibrated probability $q_i$ is generated based on the raw output (Logits) $\widehat{q}_i$ scaled as $\widehat{q}_i/T$ using a learned temperature $T$. The optimal Temperature $T$ for a trained LWALM is obtained by minimizing the Weighted Binary Cross-Entropy (BCE) loss as follows using a SOTA optimizer

$$\ell(p_i, \widehat{q}_i/T) = \epsilon^{(p_i+(1-2p_i)W)} p_i \log(\widehat{q}_i/T) + (1-p_i)\log(1-\widehat{q}_i/T)$$

where $W$ is the initialized weight.

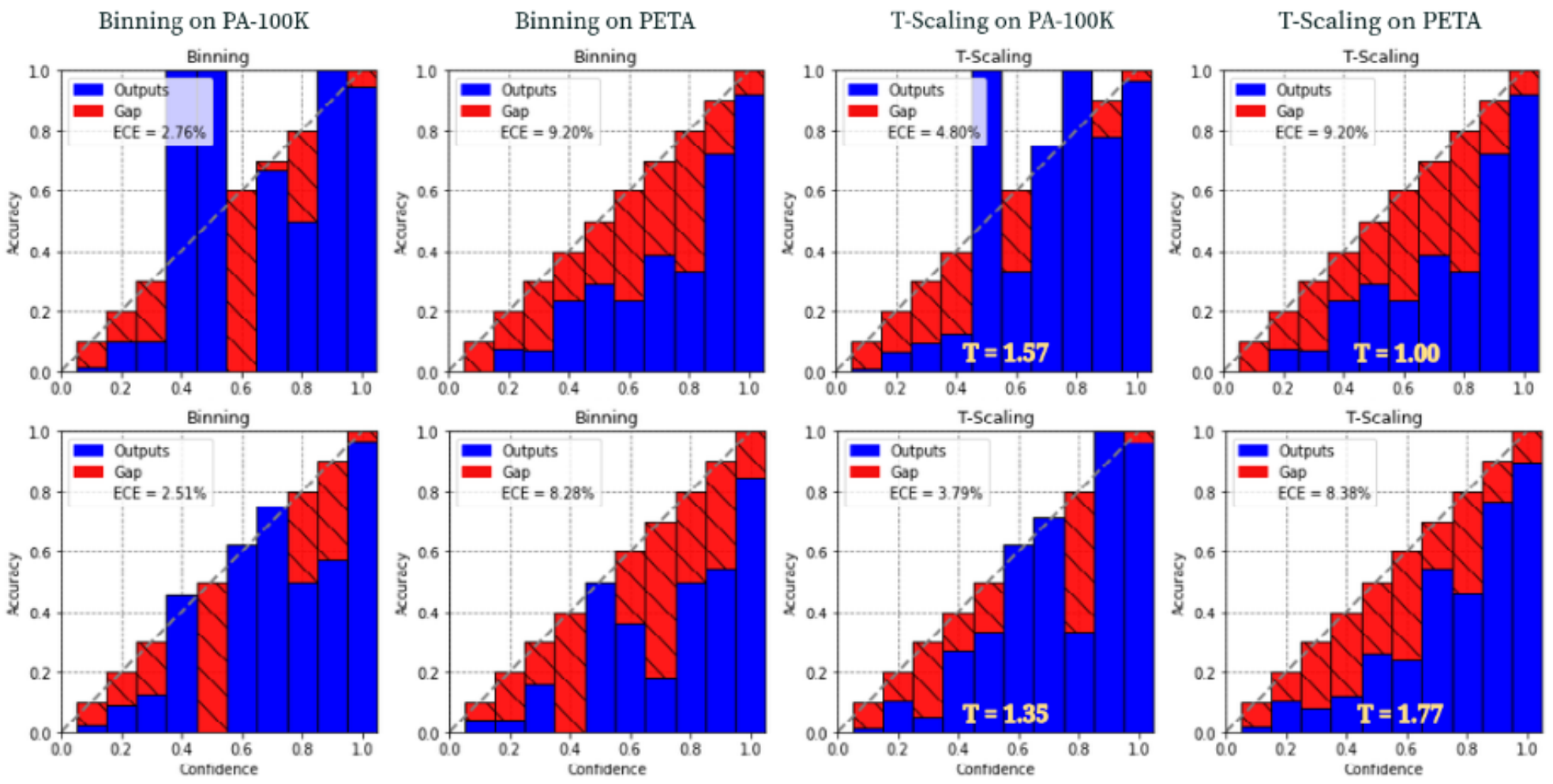

**FIGURE 2.** Reliability Diagrams for ALM (Top) and LWALM (Bottom).

We represent the calibrated model using reliability diagrams, gap metric and Expected Calibration Error (ECE) with 10 bins on both datasets (PETA and PA-100K) in Figure 2. It can be observed that ALM shows a slightly higher calibration error (2.76%, 9.20% & 4.79%, 9.29%) compared to its LW counterpart (2.51%, 8.28% & 3.79%, 8.38%) on both datasets. This improvement in LWALM is due to the optimization of the temperature scaling factor $T$, which allows the model to adjust its confidence scores more effectively. The reliability diagrams further illustrate this difference, with LWALM providing more consistent and accurate probability estimates across the range of predicted values. While both models experience some degree of miscalibration, results indicate that the integration of temperature scaling into the LWALM framework significantly enhances its ability to provide well-calibrated probability outputs. This performance boost makes LWALM a more reliable model.

## 8. Conclusion and Future works

In this paper, we present Lightweight Attribute Localization Models (LWALMs) for the Pedestrian Attribute Recognition (PAR) task. LWALMs are derived through the layer-wise compression of the Attribute Localization Model (ALM), utilizing a combination of stable CPD-EPC and SVD algorithms, along with an integrated rank search method and fine-tuning on a modified loss function. LWALMs achieve significant speedup while incurring less than a 2% accuracy drop across various pedestrian datasets. Further evaluation using reliability diagrams and metrics such as Expected Calibration Error (ECE) confirms the robustness of LWALMs, demonstrating that the model's predictive accuracy remains largely intact despite architectural and weight adjustments following compression.

While the current approach presents promising results, there remains potential to further enhance accuracy by exploring alternative optimization techniques and scaling learned parameters during training. These directions will be explored in our future work. Additionally, we aim to investigate the possibility of obtaining Lightweight models through the application of other tensor decomposition methods, such as Tensor Chain or Tensor Train decompositions.